\documentclass[10pt,twocolumn,letterpaper]{article}

\usepackage[pagenumbers]{cvpr} 

\usepackage{epsfig}

\usepackage{comment}
\usepackage{amsmath,amssymb} 
\usepackage{color}
\usepackage{cite}
\usepackage{tabularx}
\usepackage{makecell}
\usepackage{multirow}
\usepackage{booktabs}
\usepackage{graphicx}

\usepackage[pagebackref=true,breaklinks=true,letterpaper=true,colorlinks,bookmarks=false]{hyperref}

\newcommand\blfootnote[1]{%
  \begingroup
  \renewcommand\thefootnote{}\footnote{#1}%
  \addtocounter{footnote}{-1}%
  \endgroup
}

\usepackage[capitalize]{cleveref}
\crefname{section}{Sec.}{Secs.}
\Crefname{section}{Section}{Sections}
\Crefname{table}{Table}{Tables}
\crefname{table}{Tab.}{Tabs.}

\def\cvprPaperID{6605} 
\def\confName{CVPR}
\def\confYear{2023}

\begin{document}

\newcolumntype{L}[1]{>{\raggedright\arraybackslash}p{#1}}
\newcolumntype{C}[1]{>{\centering\arraybackslash}p{#1}}
\newcolumntype{R}[1]{>{\raggedleft\arraybackslash}p{#1}}

\title{BiFormer: Learning Bilateral Motion Estimation via Bilateral Transformer for 4K Video Frame Interpolation}

\author{Junheum Park\\
Korea University\\
{\tt\small jhpark@mcl.korea.ac.kr}
\and
Jintae Kim\\
Korea University\\
{\tt\small jtkim@mcl.korea.ac.kr}
\and
{Chang-Su Kim}\footnotemark\\
Korea University\\
{\tt\small changsukim@korea.ac.kr}
}

\twocolumn[{\renewcommand\twocolumn[1][]{}
\vspace*{-0.8cm}

\begin{center}
\maketitle

    \centering
    \captionsetup{type=figure}
    \vspace{-0.5cm}
    \includegraphics[width=\linewidth]{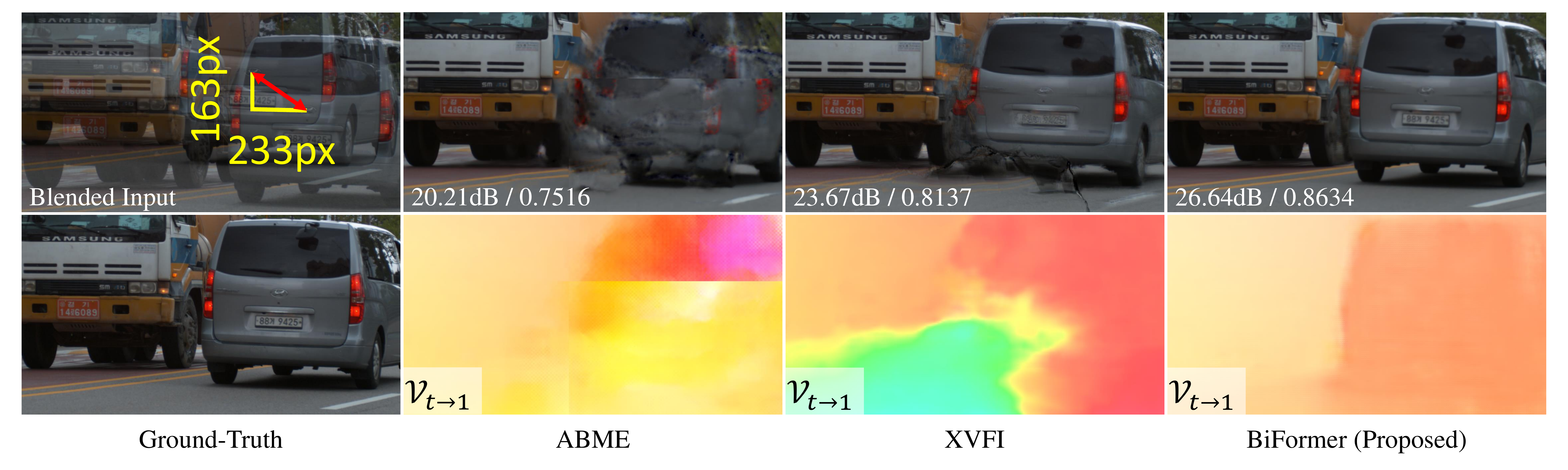}
    \vspace{-0.8cm}

    \captionof{figure}{Examples of 4K video frame interpolation results, obtained by ABME~\cite{park2021abme}, XVFI~\cite{sim2021xvfi}, and the proposed BiFormer. 4K video frame interpolation is challenging due to large motion magnitudes, \eg hundreds of pixels. PSNR/SSIM scores are presented within the interpolation results, and the estimated motion fields $\mathcal{V}_{t\rightarrow 1}$ are at the bottom row.}
\label{fig:example}
\end{center}}]


\begin{abstract}
A novel 4K video frame interpolator based on bilateral transformer (BiFormer) is proposed in this paper, which performs three steps: global motion estimation, local motion refinement, and frame synthesis. First, in global motion estimation, we predict symmetric bilateral motion fields at a coarse scale. To this end, we propose BiFormer, the first transformer-based bilateral motion estimator. Second, we refine the global motion fields efficiently using blockwise bilateral cost volumes (BBCVs). Third, we warp the input frames using the refined motion fields and blend them to synthesize an intermediate frame. Extensive experiments demonstrate that the proposed BiFormer algorithm achieves excellent interpolation performance on 4K datasets. The source codes are available at \href{https://github.com/JunHeum/BiFormer}{https://github.com/JunHeum/BiFormer}.\blfootnote{*Corresponding author.}
\end{abstract}

\vspace*{-0.2cm}
\section{Introduction}
\label{sec:intro}

Video frame interpolation (VFI) is a low-level vision task to increase the frame rate of a video, in which two (or more) successive input frames are used to interpolate intermediate frames. Its applications include video enhancement \cite{xue2019toflow}, video compression \cite{Lu2018compression, Wu2018compression}, slow-motion generation \cite{jiang2018slomo}, and view synthesis \cite{Flynn2016viewsynthesis,kalantari2016viewsynthesis}. Attempts have been made to develop effective VFI methods \cite{choi2007motion, jeong2013texture, long2016learning, liu2017dvf, niklaus2017adaconv, niklaus2017sepconv, bao2018memc, jiang2018slomo, niklaus2018ctx, liu2019cyclicgen, bao2019dain, niklaus2020softsplatting, gui2020feflow, cheng2020dsepconv, choi2020cain, lee2020adacof, park2020bmbc, park2021abme, sim2021xvfi, hu2022m2m, lu2022vfiformer, shi2022vfit}. Especially, with the advances in optical flow estimation \cite{dosovitskiy2015flownet, ilg2017flownet2, ranjan2017spynet, sun2018pwc, liu2020arflow, zhao2020maskflownet, jonschkowski2020uflow,  teed2020raft, xu2022gmflow, huang2022flowformer}, motion-based VFI methods provide remarkable performances. But, VFI for high-resolution videos, \eg 4K videos, remains challenging due to diverse factors, such as large motions and small objects, hindering accurate optical flow estimation.

Most of these VFI methods are optimized for the Vimeo90K dataset \cite{xue2019toflow} of a low spatial resolution ($448\times256$), so they tend to yield poor results on 4K videos \cite{sim2021xvfi}. It is important to develop effective VFI techniques for 4K videos, which are widely used nowadays. 4K videos are, however, difficult to interpolate, for they contain large motions as in Figure~\ref{fig:example}. To cope with large motions, many optical flow estimators adopt coarse-to-fine strategies \cite{sun2018pwc,liu2020arflow,zhao2020maskflownet}. At a coarse scale, large motions can be handled more efficiently. But, motion errors at the  coarse scale may propagate to a finer scale, making fine-scale results unreliable. To reduce such errors, the transformer can be a powerful solution, as demonstrated by recent optical flow estimators \cite{xu2022gmflow,huang2022flowformer}. However, these estimators cannot be directly used for VFI, in which the motion fields from an intermediate frame $I_t$, $0< t <1$, to input frames $I_0$ and $I_1$ should be estimated. For such bilateral motion estimation \cite{park2020bmbc, park2021abme, sim2021xvfi, huang2022rife, Reda2022film}, a novel technique is required to adopt the transformer because the source frame $I_t$ is not available.


In this paper, we propose a novel 4K VFI algorithm using the bilateral transformer (BiFormer) based on bilateral cross attention. First, we estimate global motion fields at a coarse scale via BiFormer. Second, we refine these global motion fields into final motion fields at a fine scale, by employing a motion upsampler recurrently. Last, we warp the two input frames using the final motion fields, respectively, and blend the two warped frames to synthesize an intermediate frame. Experimental results demonstrate that the proposed BiFormer algorithm provides the best performance on 4K benchmark datasets.

The work has the following major contributions:

\begin{itemize}
\itemsep0em
\item We propose the first transformer-based bilateral motion estimator, called BiFormer, for VFI.
\item We develop blockwise bilateral cost volumes (BBCVs) to refine motion fields at 4K resolution efficiently.
\item The proposed BiFormer algorithm outperforms the state-of-the-art VFI methods \cite{ding2021cdfi, niklaus2017sepconv, choi2020cain, lee2020adacof, park2020bmbc, park2021abme, sim2021xvfi} on three 4K benchmark datasets \cite{sim2021xvfi, Montgomery1994Xiph, ma2021bvidvc}.
\end{itemize}

\section{Related Work}
\label{sec:related work}

\subsection{Motion-Based VFI}

Motion-based frame warping for VFI has made great progress. An intermediate frame $I_t$, $0< t < 1$, between two input frames $I_0$ and $I_1$ can be approximated by forward warping $I_0$ with a motion field $\mathcal{V}_{0\rightarrow t}$,
\begin{equation}
\hat{I}_t = \phi_{\textrm{F}}(I_0, \mathcal{V}_{0\rightarrow t})
\end{equation}
where $\phi_{\textrm{F}}$ is the forward warping operator \cite{fant1986forwardwarping}. The required motion field $\mathcal{V}_{0\rightarrow t}$ can be obtained by scaling a motion field $\mathcal{V}_{0\rightarrow 1}$ between the input frames \cite{niklaus2018ctx,niklaus2020softsplatting,hu2022m2m}, given by
\begin{equation}
\mathcal{V}_{0\rightarrow t} = t\times \mathcal{V}_{0\rightarrow 1}.
\label{eq:v_scale}
\end{equation}
However, no motion vector in $\mathcal{V}_{0\rightarrow t}$ may pass through a certain pixel, causing a hole in the warped frame. To alleviate the hole problem, Niklaus and Liu \cite{niklaus2018ctx} predicted another warped  frame $\phi_{\textrm{F}}(I_1, \mathcal{V}_{1\rightarrow t})$, where $\mathcal{V}_{1\rightarrow t} = (1-t)\times \mathcal{V}_{1\rightarrow 0}$, and combine the two warped frames to synthesize $I_t$. On the other hand, multiple motion vectors may pass through the same pixel. To handle this collision, Niklaus and Liu \cite{niklaus2020softsplatting} introduced the softmax splatting. Hu \etal \cite{hu2022m2m} estimated reliability scores of motion vectors for better splatting.

In contrast, most motion-based VFI methods \cite{liu2017dvf,jiang2018slomo,bao2018memc,liu2019cyclicgen,bao2019dain,xue2019toflow, Reda2019ucc, park2020bmbc,park2021abme,sim2021xvfi} adopt backward warping \cite{wolberg1990invwarping},
\begin{equation}\label{Eq:b_warp}
\hat{I}_t = \phi_{\textrm{B}}(\mathcal{V}_{t\rightarrow 0},I_0).
\end{equation}
Unlike $\mathcal{V}_{0\rightarrow t}$ in \eqref{eq:v_scale}, it is not straightforward to determine the motion field $\mathcal{V}_{t\rightarrow 0}$ in \eqref{Eq:b_warp} because the intermediate frame $I_t$ is not available. Some algorithms \cite{jiang2018slomo,Reda2019ucc} assume that neighboring pixels have similar motion vectors and use motion vectors in $\mathcal{V}_{1\rightarrow 0}$ to approximate $\mathcal{V}_{t\rightarrow 0}$. Alternatively, the flow projection in \cite{bao2018memc,bao2019dain} aggregates multiple nearby motion vectors in $\mathcal{V}_{1\rightarrow 0}$ to approximate each vector in $\mathcal{V}_{t\rightarrow 0}$. To this end, Bao \etal \cite{bao2019dain} exploited depth information to determine aggregation weights adaptively. These approximate schemes, however, may cause visual artifacts near motion boundaries in warped frames.

Instead of approximation, Park \etal \cite{park2020bmbc} estimated symmetric bilateral motion vectors directly, assuming motion trajectories between $I_0$ and $I_1$ are linear. For matched pixel pairs between $I_0$ and $I_1$, symmetric bilateral motion vectors are reliable in general. However, when a pixel in $I_t$ is occluded in either $I_0$ or $I_1$, there is no matching pair, breaking the symmetry. Hence, Park \etal \cite{park2021abme} refined symmetric vectors by loosening the linear motion constraint. The resultant bilateral motion vectors become asymmetric.

Sim \etal \cite{sim2021xvfi} employed both forward and backward warping techniques. They first predicted motion fields between input frames and then forward warped these fields using themselves, $\hat{\mathcal{V}}_{t\rightarrow 0} = \phi_{\textrm{F}}(-\mathcal{V}_{0\rightarrow t},\mathcal{V}_{0\rightarrow t})$.
Then, they reconstructed $\hat{I}_t = \phi_{\textrm{B}}(\hat{\mathcal{V}}_{t\rightarrow 0}, I_0)$ via backward warping.

\subsection{Transformer}
Vaswani \etal \cite{Vaswani2017transformer} proposed the transformer based on stacked self-attention layers. With its success in NLP, the transformer has been recently employed in many vision tasks as well. Dosovitskiy \etal \cite{dosovitskiy2020vit} partitioned an image into patches and used them as tokens in the transformer. Based on global attention, the transformer is powerful but demands high complexity. To reduce the complexity, Liu \etal \cite{liu2021swin} developed the Swin transformer based on local attention and window shifting, which has been successfully adopted in dense prediction tasks.

Attempts have been made to adopt the transformer for VFI \cite{shi2022vfit,lu2022vfiformer}. Note that VFI algorithms can be classified into kernel-based or motion-based ones \cite{park2021abme}. Shi \etal \cite{shi2022vfit} extended spatial attention to spatiotemporal attention for kernel-based VFI. Lu \etal \cite{lu2022vfiformer} developed convolutional networks to determine motion fields for motion-based VFI, but they adopted the transformer for frame synthesis. In contrast, in this work, we first use the transformer to estimate bilateral motion fields for motion-based VFI.

\begin{figure*}[t]
  \centering
  \includegraphics[width=\linewidth]{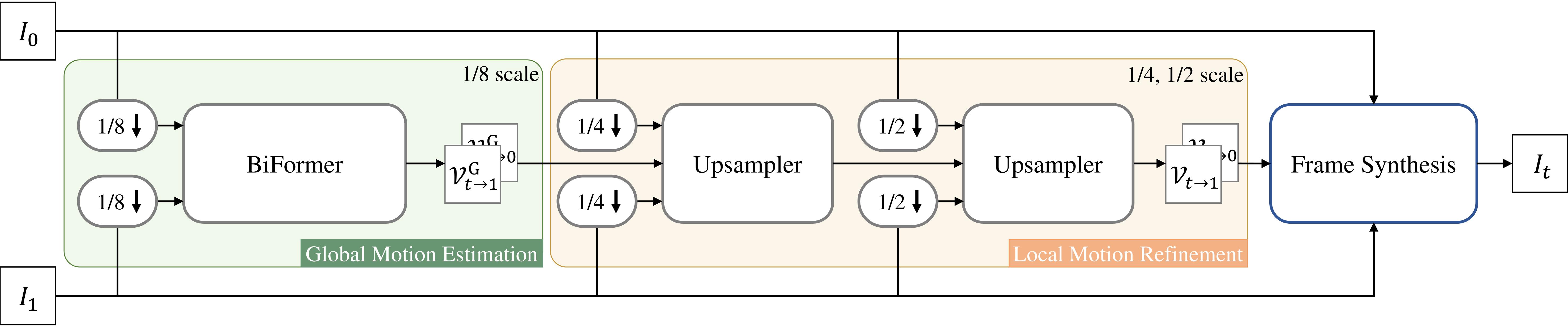}
  \vspace{-0.5cm}
  \caption{An overview of the proposed algorithm.}\label{fig:overview}
  \vspace{-0.3cm}
\end{figure*}

\begin{figure}[t]
  \centering
  \includegraphics[width=\linewidth]{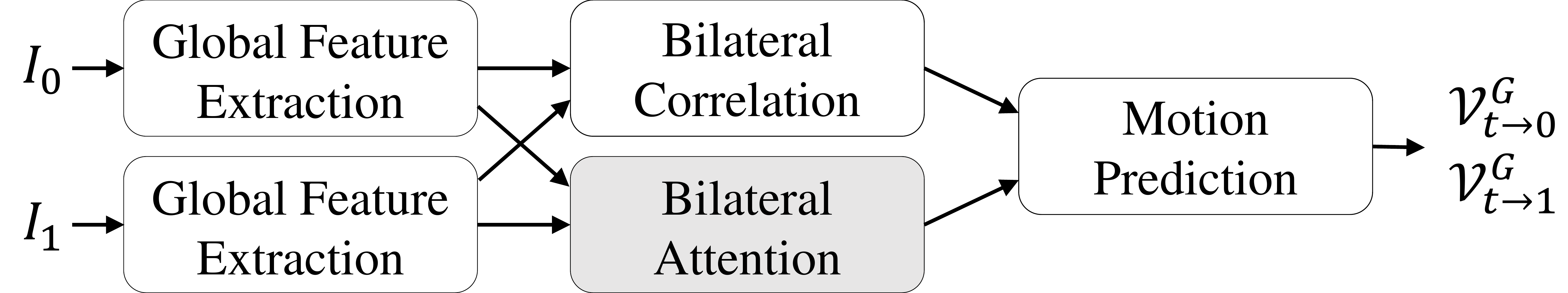}
  \vspace{-0.2cm}
  \caption{The architecture of BiFormer.}\label{fig:biformer}
  \vspace{-0.2cm}
\end{figure}

\section{Proposed Algorithm}
\label{sec:proposed algorithm}

The performance of a motion-based VFI algorithm depends on the accuracy and reliability of motion estimation. However, as shown in Figure~\ref{fig:example}, motion magnitudes are as big as hundreds of pixels in a 4K video, but such a large search window for motion vectors is impractical at the original resolution. Hence, we propose BiFormer to estimate global motion fields at a coarse scale. But, at the coarse scale, small objects and detailed texture may be lost, and their motions may be unreliable. Thus, we also develop an upsampling module to upsample and refine the global motion fields recurrently.

Figure~\ref{fig:overview} is an overview of the proposed algorithm. We first downsample input frames $I_0$ and $I_1$ and then predict global motion fields $\mathcal{V}^{\textrm{G}}_{t\rightarrow 0}$ and $\mathcal{V}^{\textrm{G}}_{t\rightarrow 1}$ via BiFormer. Then, we upsample and refine the global fields using the upsampler twice to obtain final motion fields $\mathcal{V}_{t\rightarrow 0}$ and $\mathcal{V}_{t\rightarrow 1}$. Last, we synthesize an intermediate frame $I_t$ using $\mathcal{V}_{t\rightarrow 0}$ and $\mathcal{V}_{t\rightarrow 1}$. For both $\{\mathcal{V}^{\textrm{G}}_{t\rightarrow 0}, \mathcal{V}^{\textrm{G}}_{t\rightarrow 1}\}$ and $\{\mathcal{V}_{t\rightarrow 0}, \mathcal{V}_{t\rightarrow 1}\}$, we use the symmetric bilateral motion model~\cite{park2020bmbc}.

In Figure~\ref{fig:overview}, the global estimation is performed at $1/8$ scale, while the local refinement is at $1/4$ or $1/2$ scale. Thus, input frames $I_0$ and $I_1$ are downsampled accordingly, but the downsampled images are also denoted by $I_0$ and $I_1$ for convenience. Also, it is assumed that the middle frame $I_t$, $t=\frac{1}{2}$, is interpolated from $I_0$ and $I_1$. All equations are derived for the case $t=\frac{1}{2}$, but the intermediate frame is denoted by $I_t$ for notational convenience. Note that the equations can be straigtforwardly extended for a general~$t$ ($0 < t < 1$).

\subsection{Global Motion Estimation: BiFormer}

To cope with large motions in 4K videos, transformer networks are more suitable than CNNs because of their longer-range connectivity. Hence, for global motion estimation, we propose BiFormer, which is the first transformer-based bilateral motion estimator. Figure~\ref{fig:biformer} shows the architecture of BiFormer, consisting of global feature extraction, bilateral correlation, bilateral attention, and motion prediction modules. Let us describe these modules subsequently.

\begin{figure}[t]
    \centering
    \includegraphics[width=\linewidth]{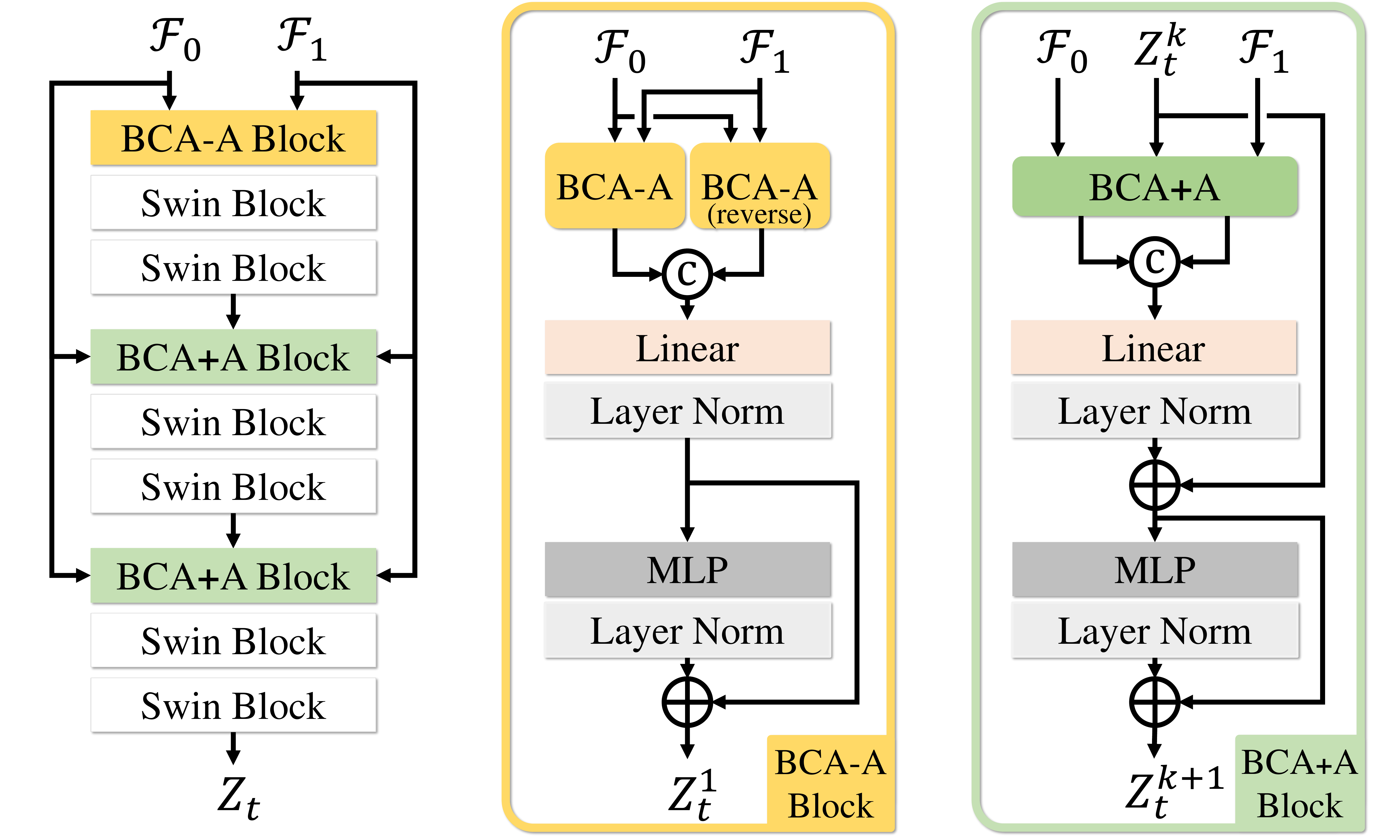}
    \caption{The architecture of the bilateral attention module.}
    \label{fig:bilateral_process}
    \vspace*{-0.2cm}
\end{figure}

\vspace*{0.1cm}
\noindent \textbf{Global feature extraction:} A transformer encoder extracts global feature maps $\mathcal{F}_0$ and $\mathcal{F}_1$ from $I_0$ and $I_1$, respectively. As the transformer encoder, we adopt the Twins architecture \cite{chu2021twins}. The encoder reduces the spatial resolution by a factor of 8. Hence, compared with the original 4K frames, $\mathcal{F}_0$ and $\mathcal{F}_1$ are at $1/64$ scale.

\begin{figure*}[t]
    \centering
    \includegraphics[width=\linewidth]{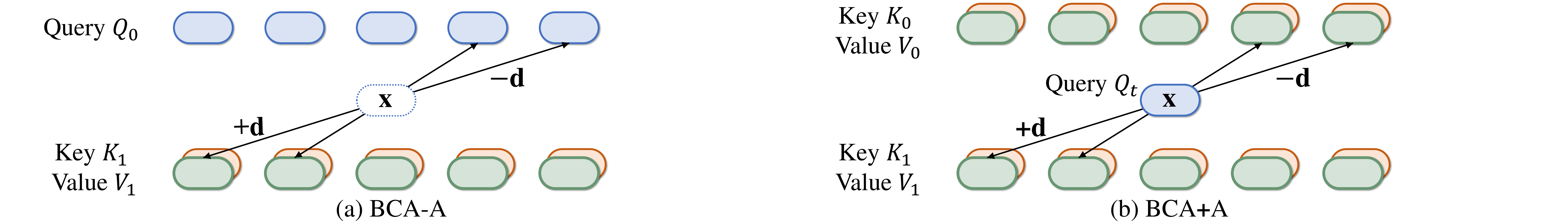}
    \vspace*{-0.5cm}
    \caption{Two types of bilateral cross attention blocks.}
    \label{fig:MBCA}
    \vspace*{-0.2cm}
\end{figure*}

\vspace*{0.1cm}
\noindent \textbf{Bilateral correlation:} Many optical flow methods compute matching costs between input frames \cite{dosovitskiy2015flownet,ilg2017flownet2,sun2018pwc,teed2020raft,park2020bmbc,park2021abme}. Similarly, we compute matching costs between the global feature maps using the bilateral correlation module \cite{park2020bmbc},
\begin{equation}\label{eq:BCV}
\mathcal{C}_t(\mathbf{x},\mathbf{d}) = \mathcal{F}_0(\mathbf{x}-\mathbf{d})^{T}\, \mathcal{F}_1(\mathbf{x}+\mathbf{d}),\; \mathbf{d} \in \mathcal{D}
\end{equation}
where $\mathcal{D} = \{\,\mathbf{d} = (d_x, d_y) \,|\, -r \leq d_x \leq r,\ -r \leq d_y \leq r\,\}$ is a local window. We set $r$ to 15, so motion magnitudes are considered up to $960$ $(=15\times 64)$ pixels vertically and horizontally at 4K resolution. In \eqref{eq:BCV}, for each pixel $\mathbf x$ in $I_t$, a symmetrically matched pair of pixels, $\mathbf{x}-\mathbf{d}$ in $I_0$ and $\mathbf{x}+\mathbf{d}$ in $I_1$, are used to compute the cost $\mathcal{C}_t(\mathbf{x},\mathbf{d})$.

\vspace*{0.1cm}
\noindent \textbf{Bilateral attention:}
In Figure~\ref{fig:bilateral_process}, the bilateral attention module consists of three types of attention blocks. First, we use a bilateral cross attention without anchor (BCA-A) block to yield $Z_t^1$, where superscript 1 means the output of the first attention block. Then, we use two Swin blocks \cite{liu2021swinv2} with the shifted windowing to yield $Z_t^3$. Next, using $Z_t^3$ as an anchor, a bilateral cross attention with anchor (BCA+A) block is used to yield $Z_t^4$. In this manner, we obtain the bilateral feature map $Z^9_t$ through nine attention blocks in total. In Figure~\ref{fig:bilateral_process}, the final output $Z^9_t$ is denoted by $Z_t$.

\vspace*{0.2cm}
\noindent
\underline{1) BCA-A:}
Cross attention \cite{Vaswani2017transformer} is used to attend two different types of features, by employing one for query and the other for key and value. In particular, in optical flow estimation, query features are extracted from a source frame, while key and value features are from a target frame \cite{xu2022gmflow}. However, in bilateral motion estimation, the source (or anchor) frame $I_t$ --- which we aim to interpolate --- is unavailable, while two target frames $I_0$ and $I_1$ are given. Thus, we develop the BCA-A block for bilateral motion estimation.

In the BCA-A block in Figure~\ref{fig:MBCA}(a), the global feature maps $\mathcal{F}_0$ and $\mathcal{F}_1$ are given. We extract query features from $\mathcal{F}_0$ and key and value features from $\mathcal{F}_1$,
\begin{align}
Q_0 &= W_{Q} \mathcal{F}_0, & K_1 &= W_{K} \mathcal{F}_1, & V_1 &= W_{V}\mathcal{F}_1,
\label{eq:QKV}
\end{align}
where $W_Q$, $W_K$, and $W_V$ are projection matrices. Then, we compute the attention matrix using a sliding window \cite{ramachandran2019stand}, given by
\begin{equation}\label{eq:MBCA}
A_t(\mathbf{x},\mathbf{d}) = Q_0(\mathbf{x}-\mathbf{d})^{T} K_1(\mathbf{x}+\mathbf{d}), \; \mathbf{d} \in \mathcal{D}
\end{equation}
With a learnable position bias $P$ \cite{liu2021swinv2}, we normalize $A_t$ by
\begin{equation}\label{Eq:softmax}
\bar{A}_t = \textrm{SoftMax}(A_t + P),
\end{equation}
so that $\sum_{\mathbf{d}\in \mathcal{D}}{\bar{A}_t}(\mathbf{x},\mathbf{d})=1$ for each $\mathbf{x}$. Thus, $\bar{A}_t(\mathbf{x},\mathbf{d})$ represents the similarity between $\mathbf{x}-\mathbf{d}$ in $I_0$ and $\mathbf{x}-\mathbf{d}$ in $I_1$, which are symmetrically located with respect to $\mathbf{x}$ in $I_t$ as shown in Figure~\ref{fig:MBCA}(a). Finally, based on the similarities, we obtain the attended value feature,
\begin{equation}
\textstyle
Z_{1\rightarrow t}(\mathbf{x}) = \sum_{\mathbf{d}\in \mathcal{D}}{\bar{A}_t(\mathbf{x},\mathbf{d})^{T}\,{V}_1(\mathbf{x}+\mathbf{d})}.
\end{equation}
Note that $Z_{1\rightarrow t}$ uses the information in $I_1$ to approximately represent $I_t$ based on the symmetric bilateral matching information in \eqref{eq:MBCA}.

Similarly, we reverse the roles of $\mathcal{F}_0$ and $\mathcal{F}_1$ in \eqref{eq:QKV} and then perform the same BCA-A process to obtain another attended value feature $Z_{0\rightarrow t}$. Then, as shown in Figure~\ref{fig:bilateral_process}, we concatenate $Z_{0\rightarrow t}$ and $Z_{1\rightarrow t}$ and process the result through linear, layer normalization, and MLP layers to yield $Z_t^1$.

\vspace*{0.2cm}
\noindent
\underline{2) BCA+A:} After the first BCA-A block, the anchor information $Z_t^{k}$ at time $t$ is available. We can aggregate more informative features for $I_t$ by exploiting $Z_t^{k}$. As shown in Figure~\ref{fig:MBCA}(b), if pixel $\mathbf{x}$ in $Z_t^{k}$ has a constant velocity, its trajectory formed by $\mathbf{x}-\mathbf{d}$ in $\mathcal{F}_0$, $\mathbf{x}$ in $Z_t^k$, and $\mathbf{x}+\mathbf{d}$ in $\mathcal{F}_1$ is linear. Hence, we develop the BCA+A block to exploit mutual connections among the triplet ($\mathbf{x}-\mathbf{d}$, $\mathbf{x}$, $\mathbf{x}+\mathbf{d}$).

First, we extract queries from the anchor $Z_t^k$ and keys and values from the two target feature maps $\mathcal{F}_0$ and $\mathcal{F}_1$,
\begin{equation}
\begin{aligned}
Q_t &= W_{Q} Z_t^k, & K_0 &= W_{K} \mathcal{F}_0, & V_0 &= W_{V}\mathcal{F}_0, \\
    &               & K_1 &= W_{K} \mathcal{F}_1, & V_1 &= W_{V}\mathcal{F}_1.
\end{aligned}
\end{equation}
Then, we compute the anchor-aware attention matrix,
\begin{equation}
B_t(\mathbf{x},\mathbf{d}) = Q_t(\mathbf{x})^{T} K_0(\mathbf{x}-\mathbf{d}) + Q_t(\mathbf{x})^{T} K_1(\mathbf{x}+\mathbf{d})
\end{equation}
where $\mathbf{d} \in \mathcal{D}$. We convert $B_t$ into $\bar{B}_t$ similarly to \eqref{Eq:softmax}. Finally, we obtain the anchor-aware attended value features,
\begin{align}
Z^{\textrm{anch}}_{0\rightarrow t} (\mathbf{x}) &= \textstyle \sum_{\mathbf{d}\in \mathcal{D}}{\bar{B}_t(\mathbf{x},\mathbf{d})^{T} V_0(\mathbf{x}-\mathbf{d})}, \\
Z^{\textrm{anch}}_{1\rightarrow t} (\mathbf{x}) &= \textstyle \sum_{\mathbf{d}\in \mathcal{D}}{\bar{B}_t(\mathbf{x},\mathbf{d})^{T} V_1(\mathbf{x}+\mathbf{d})}.
\end{align}
As done in the BCA-A block, these two features are concatenated and then processed to yield $Z_t^{k+1}$.

\vspace*{0.2cm}
\noindent \textbf{Motion prediction:} To predict global motion fields $\mathcal{V}^{\textrm{G}}_{t\rightarrow 0}$ and $\mathcal{V}^{\textrm{G}}_{t\rightarrow 1}$, we concatenate the cost volume $\mathcal{C}_t$ in \eqref{eq:BCV} and the bilateral feature map $Z_t^9$ from the bilateral attention module. Then, the concatenated features are processed by convolution layers to yield the global bilateral motion field ${\cal V}^{\rm G}_{t\rightarrow 1}$ from $I_t$ to $I_1$. Also, because of the symmetric bilateral motion constraint \cite{park2020bmbc}, we have
\begin{equation}
{\cal V}^{\rm G}_{t \rightarrow 0} = - {\cal V}^{\rm G}_{t \rightarrow 1}.
\label{eq:symmetry_condition}
\end{equation}
More details are presented in the supplement.

\begin{figure*}[t]
    \centering
    \includegraphics[width=\linewidth]{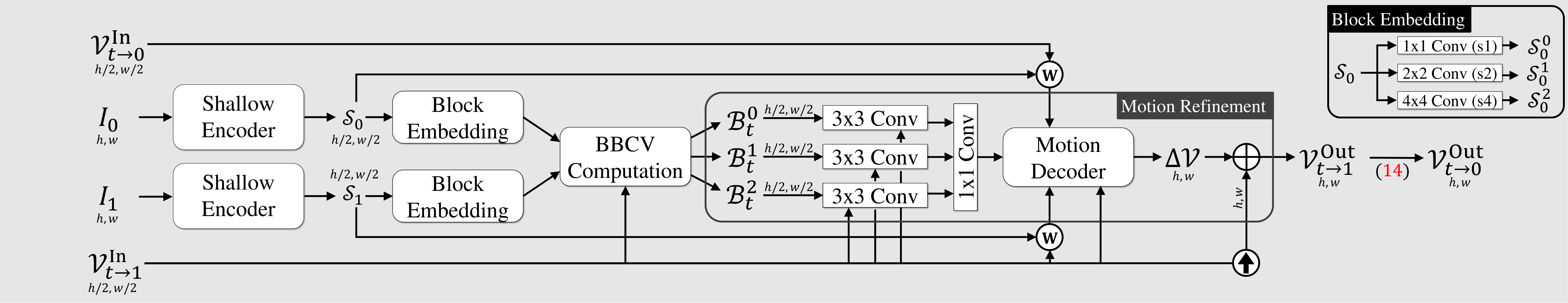}
    \caption{The network structure of the upsampler. In the block embedding layer, (s$N$) denotes that the convolution layer has stride $N$.}
    \label{fig:LocalRefinement}
    \vspace{-0.1cm}
\end{figure*}

\subsection{Local Motion Refinement: Upsampler}

At 1/8 scale, BiFormer predicts large global motions such as camera panning and rigid movements of large objects effectively. However, the global motion fields may be unreliable, especially in regions for small objects or near sharp object boundaries. Hence, we develop the upsampler to refine the global motion fields. As shown in Figure~\ref{fig:overview}, we use the upsampler twice to refine the global motion fields ${\cal V}^{\rm G}_{t \rightarrow 1}$ and ${\cal V}^{\rm G}_{t \rightarrow 0}$ at 1/8 scale into the final motion fields ${\cal V}_{t \rightarrow 1}$ and ${\cal V}_{t \rightarrow 0}$ at 1/2 scale. Since the upsampler operates at the fine scales, we implement it based on convolution layers, instead of transformer blocks. This is because the transformer demands a huge number of parameters to process high-resolution input in general.

Figure~\ref{fig:LocalRefinement} shows the network structure of the upsampler, which takes a motion field ${\cal V}_{t \rightarrow 1}^{\rm In}$ as input and yields an upsampled field ${\cal V}_{t \rightarrow 1}^{\rm Out}$ as output. Note that
\begin{equation}
\textstyle
{\cal V}_{t \rightarrow 0}^{\rm In} = -{\cal V}_{t \rightarrow 1}^{\rm In} \quad \mbox{ and } \quad {\cal V}_{t \rightarrow 0}^{\rm Out} = -{\cal V}_{t \rightarrow 1}^{\rm Out}.
\end{equation}

\vspace*{0.2cm}
\noindent \textbf{Feature extraction:}
Given $I_0$ and $I_1$, we extract the feature maps using the shallow encoder, which yields local feature maps $\mathcal{S}_0$ and $\mathcal{S}_1$. Note that the local motion refinement is performed at fine scales of 1/4 and 1/2. Thus, to refine large motions, the search range should be sufficiently large as well. We hence perform block embedding using three convolution layers with kernel sizes of $1\times1$, $2\times2$, and $4\times4$, respectively. They process $\mathcal{S}_0$ and $\mathcal{S}_1$ to generate $\mathcal{S}^k_0$ and $\mathcal{S}^k_1$, where $k \in \{0, 1, 2 \}$ is the block size index. Note that the block embedding layer is depicted in detail at the top right corner of Figure~\ref{fig:LocalRefinement}.

\begin{figure}[t]
    \centering
    \includegraphics[width=0.9\linewidth]{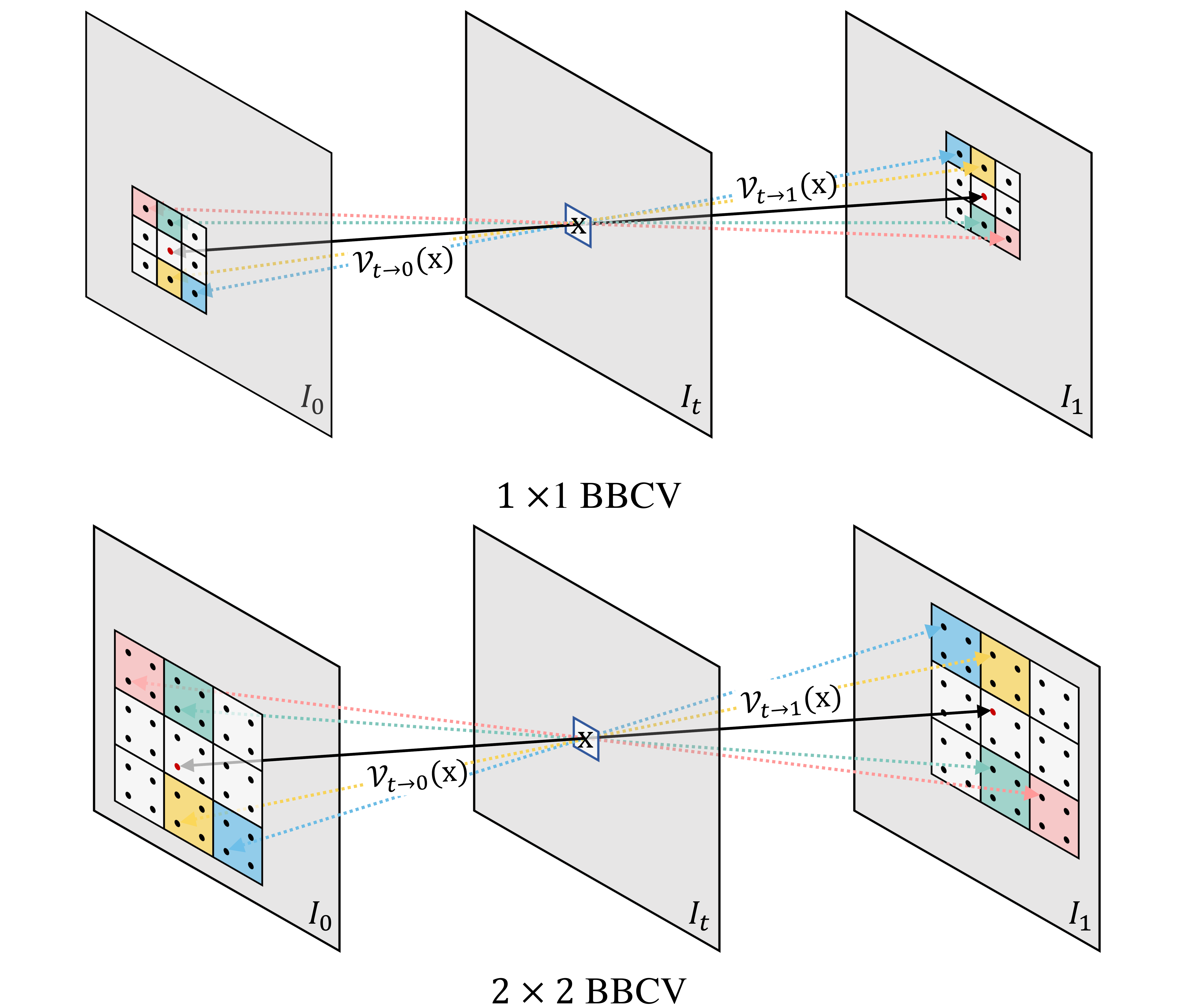}
    \caption{Illustration of a $1\times1$ BBCV ${\cal B}_t^0$ and a $2\times2$ BBCV ${\cal B}_t^1$.}
    \label{fig:BBCV}
    \vspace{-0.1cm}
\end{figure}

\begin{table*}[t]
    \caption
  {
        Quantitative comparison of VFI results. For each test, the best result is {\textbf{boldfaced}}, while the second-best is \underline{underlined}.
    }
    \vspace*{-0.1cm}
    \centering
    {\footnotesize
    \begin{tabular}{L{2.8cm}C{0.9cm}C{0.9cm}C{0.9cm}C{0.9cm}C{0.9cm}C{0.9cm}C{1.5cm}C{1.5cm}}
    \toprule
    \multirow{2}[2]{*}{} &\multicolumn{2}{c}{X4K1000FPS} & \multicolumn{2}{c}{Xiph-4K} & \multicolumn{2}{c}{BVI-DVC-4K} & \multirow{2}[2]{*}{{\makecell{2K Runtime \\ (seconds)}}} & \multirow{2}[2]{*}{{\makecell{\#Parameters \\ (millions)}}} \\
    \cmidrule(lr){2-3} \cmidrule(lr){4-5} \cmidrule(lr){6-7}
    & PSNR& SSIM& PSNR& SSIM&PSNR &SSIM & & \\
    \midrule
    SepConv\cite{niklaus2017sepconv} &24.34&0.7420& 32.61&0.8801 & 26.57 & 0.8485 & 0.41 & 21.6\\[0.15em]
    CAIN\cite{choi2020cain} & 24.50 & 0.7522& 33.07 & 0.8896& 27.06 & 0.8582 & 0.14 & 42.8\\[0.15em]
    AdaCoF\cite{lee2020adacof}& 24.13&0.7338& 32.72 &0.8805& 26.35 & 0.8402 & 0.28 &  22.9\\[0.15em]
    BMBC\cite{park2020bmbc} &22.86 &0.7269 & 31.27& 0.8804& 25.41 & 0.8384 & 6.45 &  11.0\\[0.15em]
    CDFI\cite{ding2021cdfi} & 24.49&0.7419& 33.01&0.8720& 26.84 & 0.8496 & 1.24 & 5.0\\[0.15em]
    XVFI\cite{sim2021xvfi} & 30.12&0.8704& 34.06&0.8946& 29.17& 0.8956 & 0.36 & 5.6\\[0.15em]
    ABME\cite{park2021abme} & 30.16&0.8793& 33.81&0.9030& 28.28 & 0.8905 & 1.16 & 18.1\\[0.15em]
    VFIformer\cite{lu2022vfiformer} & 24.58 &0.8054& 33.69 & 0.9252 &  27.45 & 0.9049 & -- & 24.2 \\[0.15em]
    M2M-PWC\cite{hu2022m2m} & {\underline{30.81}}&{\underline{0.9120}}&{\underline{34.46}} &{\underline{0.9252}}& {\textbf{29.77}} & {\underline{0.9274}} & 0.07 & 7.6\\[0.15em]
    BiFormer {\footnotesize (Proposed)} & {\textbf{31.32}}& {\textbf{0.9212}}& {\textbf{34.48}}& {\textbf{0.9268}}& {\underline{29.67}}& {\textbf{0.9296}}& 0.53 & 11.2 \\
    \bottomrule\\[-2em]
    \end{tabular}
    }
    \label{table:Evaluation on test set}
\end{table*}

\vspace*{0.2cm}
\noindent \textbf{BBCVs:}
In the optical flow estimator in \cite{teed2020raft}, blockwise cost volumes are used to increase the search range. Those volumes, however, cannot be used in this work, since the source frame $I_t$ is unavailable and should be interpolated. We hence develop BBCVs.

Figure~\ref{fig:BBCV} illustrates BBCVs. Given the bilateral motion fields $\mathcal{V}_{t\rightarrow 0}^{\rm In}$ and $\mathcal{V}_{t\rightarrow 1}^{\rm In}$, which are symmetric with respect to $I_t$, the bilateral search windows for pixel $\mathbf{x}$ have two center points (red points in Figure~\ref{fig:BBCV}), given by
\begin{equation}
\begin{aligned}
\mathbf{x}'^k_0 &= (\mathbf{x} + \mathcal{V}^{\rm In}_{t\rightarrow 0}(\mathbf{x}))/2^k,\\
\mathbf{x}'^k_1 &= (\mathbf{x} + \mathcal{V}^{\rm In}_{t\rightarrow 1}(\mathbf{x}))/2^k,
\end{aligned}
\end{equation}
where $k$ is the block size index. Then, we define a search window $\{\mathbf{d} = (d_x, d_y) \,|\, -r \leq d_x \leq r,\ -r \leq d_y \leq r\,\}$ with $r=2$ and compute three BBCVs $\{\mathcal{B}^0_t, \mathcal{B}^1_t, \mathcal{B}^2_t\}$,
\begin{equation}\label{Eq:bbcv}
\mathcal{B}^k_t(\mathbf{x}, \mathbf{d}) = {\mathcal{S}^k_0(\mathbf{x}'^k_0-  \mathbf{d})}^{T} \mathcal{S}^k_1(\mathbf{x}'^k_1+\mathbf{d}),  \,\, k=0, 1, 2.
\end{equation}
Since the source pixel $\mathbf{x}$ in $\mathcal{B}^k_t$ in \eqref{Eq:bbcv} is at the high resolution, the motion of even a small object can be predicted precisely. On the other hand, the size of the search window is $(2r+1)^2$ blocks, which corresponds to $((2r+1)\times2^k)^2$ in the pixel unit. Thus, a large motion can be refined by employing BBCVs.

\vspace*{0.2cm}
\noindent \textbf{Motion refinement:}
For each $k \in \{0,1,2\}$, the BBCV $\mathcal{B}^k_t$ and the motion field $\mathcal{V}_{t\rightarrow 1}^{\rm In}$ are concatenated and processed by a $3\times 3$ convolution layer. Then, the three results are aggregated by a $1\times1$ convolution layer, yielding the matching feature map.

Using $\mathcal{V}_{t\rightarrow 0}^{\rm In}$ and $\mathcal{V}_{t\rightarrow 1}^{\rm In}$, we warp the local feature maps $\mathcal{S}_0$ and $\mathcal{S}_1$. Then, by taking the warped feature maps, the matching feature map, and $\mathcal{V}_{t\rightarrow 1}^{\rm In}$ as input, the motion decoder predicts a residual motion field $\Delta \mathcal{V}$. Finally, we generate an upsampled, refined motion field given by
\begin{equation}
\mathcal{V}^{\rm Out}_{t\rightarrow 1} = \tilde{\mathcal{V}}^{\rm In}_{t\rightarrow 1} + \Delta \mathcal{V}
\end{equation}
where $\tilde{\mathcal{V}}^{\rm In}_{t\rightarrow 1}$ is the bilinearly upsampled $\mathcal{V}_{t\rightarrow 1}^{\rm In}$.

\subsection{Frame Synthesis}
We develop a simple frame synthesis network, composed of an encoder, three skip connections with warping, and a decoder. The encoder processes $I_0$ and $I_1$ to extract multi-scale feature maps $\mathcal{G}^l_0$ and $\mathcal{G}^l_1$, where $l\in\{0,1,2\}$ is the scale index. At level $l$, we downsample the refined motion fields $\mathcal{V}_{t\rightarrow 0}$ and $\mathcal{V}_{t\rightarrow 1}$ to yield $\mathcal{V}^l_{t\rightarrow 0}$ and $\mathcal{V}^l_{t\rightarrow 1}$. Using them, we warp $\mathcal{G}^l_0$ and $\mathcal{G}^l_1$, which are then passed to the decoder via a skip connection to the first layer of the decoder at the $l$th level. Finally, the decoder synthesizes the intermediate frame $I_t$. The details are presented in the supplement.

\section{Experiments}

\subsection{Training}
We first train the global motion estimator, BiFormer. Then, we freeze BiFormer and train the local motion refinement network and the synthesis network together.

\vspace*{0.2cm}
\noindent \textbf{BiFormer:}
To train BiFormer, we define the photometric loss between the ground-truth $I^{\textrm{GT}}_{t}$ and two warped frames:
\begin{align}
    \mathcal{L}_\textrm{pho} &=  \rho(I^{\textrm{GT}}_{t}-\phi_\textrm{B}(\mathcal{V}^{\rm G}_{t\rightarrow 0}, I_0)) + \rho(I^{\textrm{GT}}_{t}-\phi_\textrm{B}(\mathcal{V}^{\rm G}_{t\rightarrow 1}, I_1)) \nonumber\\
    + & \mathcal{L}_\textrm{cen}(I^\textrm{GT}_t,\phi_\textrm{B}(\mathcal{V}^{\rm G}_{t\rightarrow 0}, I_0)) + \mathcal{L}_\textrm{cen}(I^\textrm{GT}_t,\phi_\textrm{B}(\mathcal{V}^{\rm G}_{t\rightarrow 1}, I_1))
    \label{Eq:photometric_loss}
\end{align}
where $\rho(x)=(x^2+\epsilon^2)^{\alpha}$ is the Charbonnier function \cite{Char1994loss} and $\mathcal{L}_\textrm{cen}$ is the census loss \cite{Meister2018unflow, zhong2019unsupervised, zou2018DF}. The parameters are set to $\alpha=0.5$ and $\epsilon=10^{-3}$. We use only the Vimeo90K training set \cite{xue2019toflow} to train BiFormer. It is composed of 51,312 triplets of resolution $448\times256$. Thus, in inference, for the global motion estimation, we downsample input frames to $448\times256$.

\begin{figure*}[t]
    \centering
    \includegraphics[width=\linewidth]{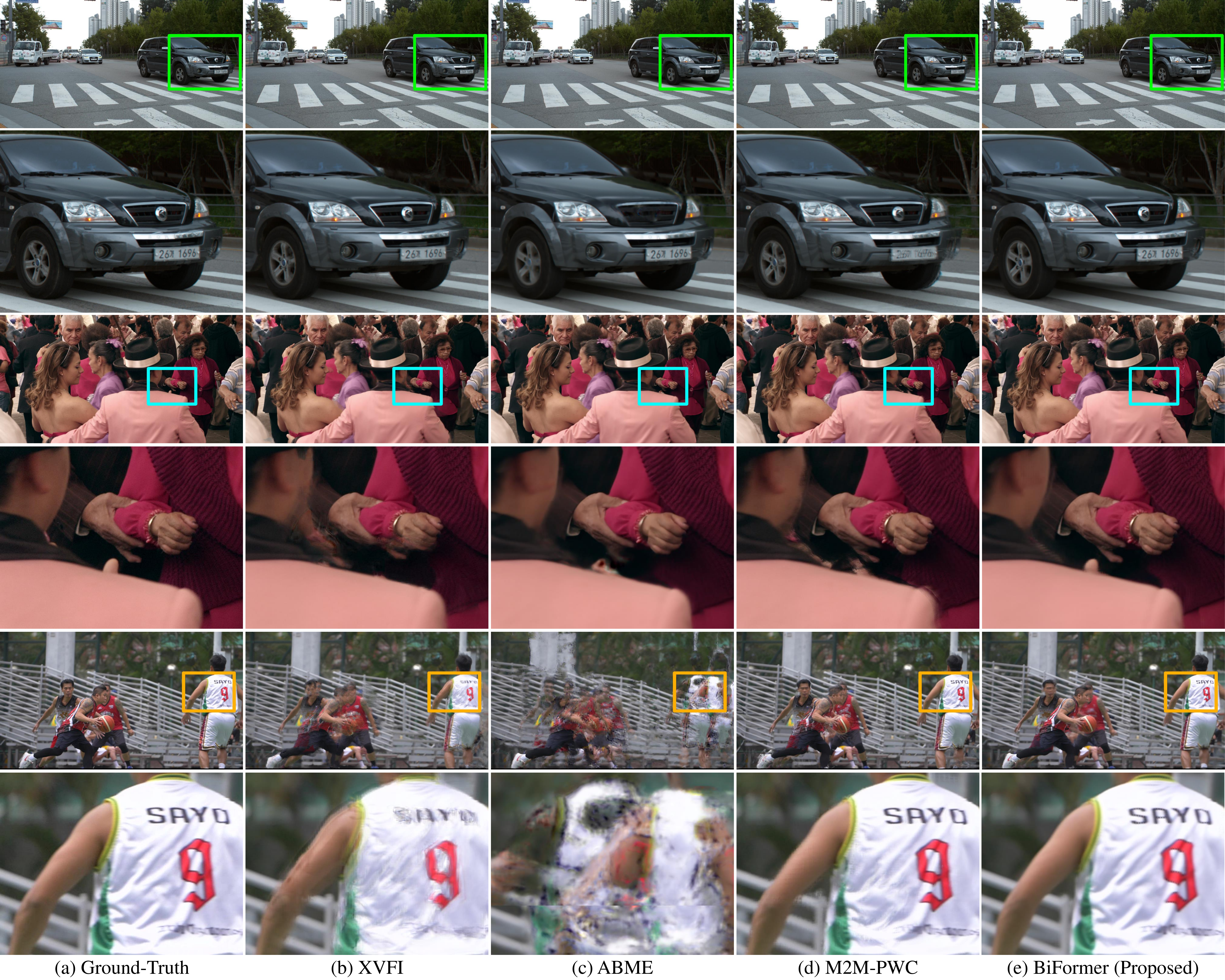}
    \vspace*{-0.5cm}
    \caption{
    Qualitative comparison of interpolated frames. The top, middle, and bottom examples are from X4K1000FPS, Xiph-4K, and BVI-DVC-4K, respectively. The proposed BiFormer in (e) interpolates the frames faithfully to the ground truth in (a).}
    \label{fig:Eval on X4K1000FPS}
    \vspace*{-0.2cm}
\end{figure*}

\vspace*{0.2cm}
\noindent \textbf{Local motion refinement with synthesis:}
To train the local motion refinement network and the synthesis network together, we define the synthesis loss between the ground-truth $I^{\textrm{GT}}_{t}$ and a synthesized frame $I_t$ as
\begin{equation}
    \label{eq:synthesis loss}
    \textstyle
    \mathcal{L}_\textrm{syn} = \rho(I^\textrm{GT}_t-I_t) + \mathcal{L}_\textrm{cen}(I^\textrm{GT}_t,I_t).
\end{equation}
We use the training dataset of X4K1000FPS \cite{sim2021xvfi}, containing 4,408 sets of 65 successive frames of resolution $768\times768$, cropped from 4K videos. It is augmented by random flipping, rotating, order reversing, and cropping of $512\times512$ patches. Moreover, to learn the motion refinement at various resolutions, we downsample input frames to a random size from $96\times96$ to $256\times256$. We use the Adam optimizer \cite{kingma2014adam} with a learning rate of $\eta = 10^{-4}$ until 0.1M iterations and then halve $\eta$ after every 0.05M iterations. We use a batch size of 4 for 0.2M iterations in total.

\subsection{Test Datasets}

As mentioned above, we use the Vimeo90K training set for BiFormer and then the X4K1000FPS training set for the whole algorithm. The proposed algorithm is designed for 4K videos, so we use three 4K test datasets.

\vspace*{0.2cm}
\noindent \textbf{X4K1000FPS \cite{sim2021xvfi}:}
Its test set, called X-TEST, contains 15 clips of 33 successive 4K frames of frame rate 1000 fps. These clips contain diverse motions, including rotation, panning, rigid movement, zoom-in, and zoom-out.

\vspace*{0.2cm}
\noindent \textbf{Xiph-4K \cite{Montgomery1994Xiph}:}
It contains 19 raw video sequences for testing video codecs. Each sequence is composed of 31 successive 4K frames. Even-indexed frames are used as input frames, while odd-indexed frames are the ground-truth for intermediate frames. Thus, there are 285 triplets in total.

\vspace*{0.2cm}
\noindent \textbf{BVI-DVC-4K \cite{ma2021bvidvc}:}
It contains 200 raw sequences. From each sequence, three successive 4K frames are extracted. Thus, there are 200 triplets in total. These sequences capture complex texture and cluttered scenes.

\subsection{Comparison with the State-of-the-Arts}

We compare the proposed BiFormer algorithm with nine conventional algorithms: SepConv \cite{niklaus2017sepconv}, CAIN \cite{choi2020cain}, AdaCoF \cite{lee2020adacof}, BMBC \cite{park2020bmbc}, CDFI \cite{ding2021cdfi}, XVFI \cite{sim2021xvfi}, ABME \cite{park2021abme}, VFIformer \cite{lu2022vfiformer}, and M2M-PWC \cite{hu2022m2m}. Table~\ref{table:Evaluation on test set} compares the average PSNR and SSIM scores on X4K1000FPS, Xiph-4K, and BVI-DVC-4K. Except for the PSNR metric on BVI-DVC-4K, BiFormer provides the best PSNR and SSIM scores. Especially, on X4K1000FPS, compared to the second-best M2M-PWC, BiFormer improves PSNR by more than 0.5dB. It is worth pointing out that, whereas M2M-PWC uses PWC-Net trained on an optical flow dataset, BiFormer achieves better results without employing such additional datasets.

Table \ref{table:Evaluation on test set} also lists the runtimes for interpolating an intermediate frame in a 2K ($1920\times1080$) sequence using an RTX 3090 GPU. Note that, for a 4K sequence, BMBC, CDFI, VFIformer, and ABME cannot process entire frames at once due to the lack of memory. Thus, they should divide input frames into patches, infer an intermediate frame patch-wise, and merge the interpolated patches. This patch-wise processing increases runtimes significantly. On the contrary, BiFormer can interpolate a 4K frame at once. Thus, for a fair comparison with the conventional methods, we compare the runtimes at 2K resolution, instead of 4K resolution.

Figure~\ref{fig:Eval on X4K1000FPS} shows interpolation results qualitatively. Due to large motions in the 4K frames, the conventional algorithms fail to predict the fingers in the middle image and the player in the bottom image properly, yielding ghost and deformation artifacts. Moreover, they cannot handle the fast motions of a car in the top image, causing blurry artifacts and missing object parts. In contrast, BiFormer reconstructs them more faithfully without noticeable artifacts.

\subsection{Analysis}

Let us analyze the VFI performance of the proposed algorithm on X4K1000FPS.

\vspace*{0.2cm}
\noindent \textbf{Bilateral process}:
In the bilateral attention module in Figure~\ref{fig:bilateral_process}, there are six Swin blocks and three BCA blocks (one BCA-A block and two BCA+A blocks).
Table \ref{table:ablation_global} lists the performances when the BCA-A or BCA+A blocks are removed while the six Swin blocks are maintained. We see that the performance improves as more BCA blocks are used, which confirms the effectiveness of the BCA blocks.

\begin{table}[t]
    \caption
    {
        Ablation studies for the global motion estimation.
    }
    \vspace*{-0.2cm}
    \centering
    {\footnotesize
    \begin{tabular}{C{0.6cm}C{1cm}C{1.2cm}C{1.2cm}C{0.9cm}C{0.9cm}}
    \toprule
    \multirow{2}[1]{*}{\makecell{6 swin}}& \multicolumn{3}{c}{BCA block} & \multicolumn{2}{c}{X4K1000FPS} \\
    \cmidrule(lr){2-4} \cmidrule(lr){5-6}
    & BCA-A & BCA+A$^{\textrm{1st}}$ & BCA+A$^{\textrm{2nd}}$  & PSNR& SSIM\\
    \midrule
    \checkmark & & & & 30.94 & 0.8867\\
    \checkmark &\checkmark  & & & 31.20 & 0.9086\\
    \checkmark &\checkmark  &\checkmark  & & 31.26 & 0.9158\\
    \checkmark &\checkmark  &\checkmark  &\checkmark &31.32 & 0.9212\\
    \bottomrule
    \end{tabular} \\
    }
    \label{table:ablation_global}
    \vspace*{-0.15cm}
\end{table}

\begin{table}[t]
    \caption
    {
        Ablation studies for the local motion refinement.
    }
    \vspace*{-0.2cm}
    \footnotesize
    \centering
    \begin{tabular}{L{2.9cm}L{1.9cm}C{0.9cm}C{0.9cm}}
    \toprule
    \multirow{2}[2]{*}{}&\multirow{2}[2]{*}{Setting}&\multicolumn{2}{c}{X4K1000FPS}\\
     \cmidrule(lr){3-4}
     &  & PSNR & SSIM\\[-0.1em]
    \midrule
    \multirow{2}[1]{*}{Search range $r$}& 2 (default) & 31.32 & 0.9212\\
     & 1 & 31.20 & 0.9037\\
    \midrule
    \multirow{3}[1]{*}{Maximum block size}& $4\times4$ (default)
     & 31.32 & 0.9212\\
     & $2\times2$ & 31.27 & 0.9161\\
     & $1\times1$ & 31.16 & 0.8914\\
    \midrule
    \multirow{3}[1]{*}{Refinement scales}& 1/4, 1/2 (default)
     & 31.32 & 0.9212\\
     & 1/2 & 31.15 & 0.8895\\
     & 1/4 & 30.88 & 0.8859\\
    \bottomrule\\[-1.5em]
    \end{tabular}
    \label{table:ablation_local}
\end{table}

\noindent \textbf{Local motion refinement}:
Table \ref{table:ablation_local} analyzes the impacts of two hyper-parameters of BBCVs in the local motion refinement: the size $r$ of the search range and the maximum block size. First, the performance degrades when $r$ is smaller than the default value 2. A bigger $r$ (\eg 4) increases memory complexity and makes it impossible to interpolate a 4K frame at once. Second, the maximum block size $4\times4$ means that the three BBCVs for $1\times1$, $2\times2$, and $4\times4$ block sizes are employed. The performance improves as more BBCVs are used, for a bigger block size enables the cost volume to consider larger motions more efficiently.

Table \ref{table:ablation_local} also analyzes the impacts of refinement scales. In the default mode, we use the upsampler twice to refine the motion fields at $1/4$ and $1/2$ scales sequentially. If we refine them at the coarse scale of $1/4$ only, the performance degrades severely because fine motions are less accurately represented. If we instead do the refinement at the fine scale of $1/2$ directly, the performance gets better but is worse than the sequential refinement. This is because the sequential refinement takes advantage of both the reliability at the coarse scale and the accuracy at the fine scale.

\begin{figure}[t]
    \centering
    \includegraphics[width=\linewidth]{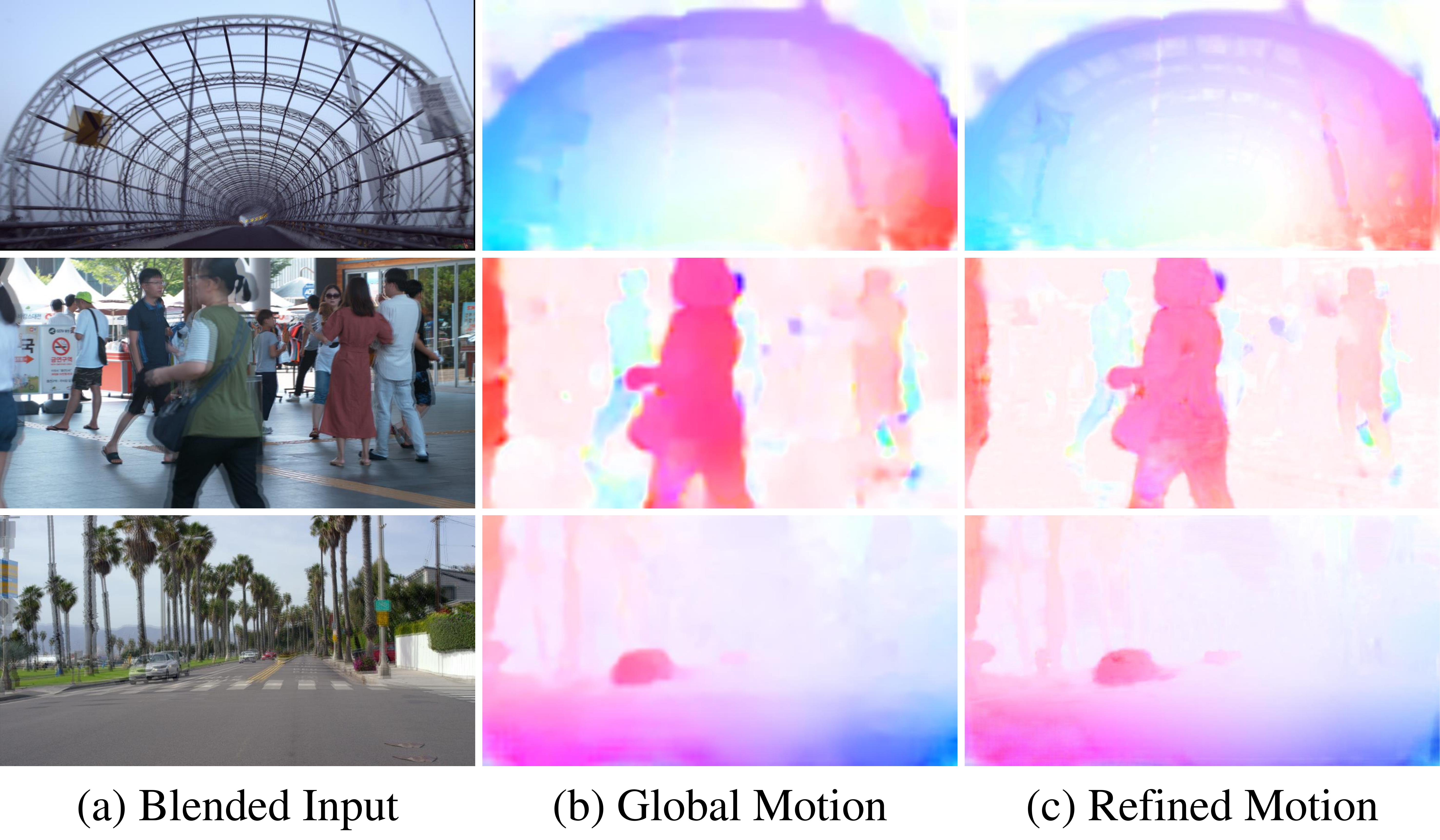}
    \vspace*{-0.6cm}
    \caption{
    Visualization of global and refined motion fields.}
    \label{fig:motion_comparison}
    \vspace*{-0.2cm}
\end{figure}

\begin{figure}[t]
    \centering
    \includegraphics[width=\linewidth]{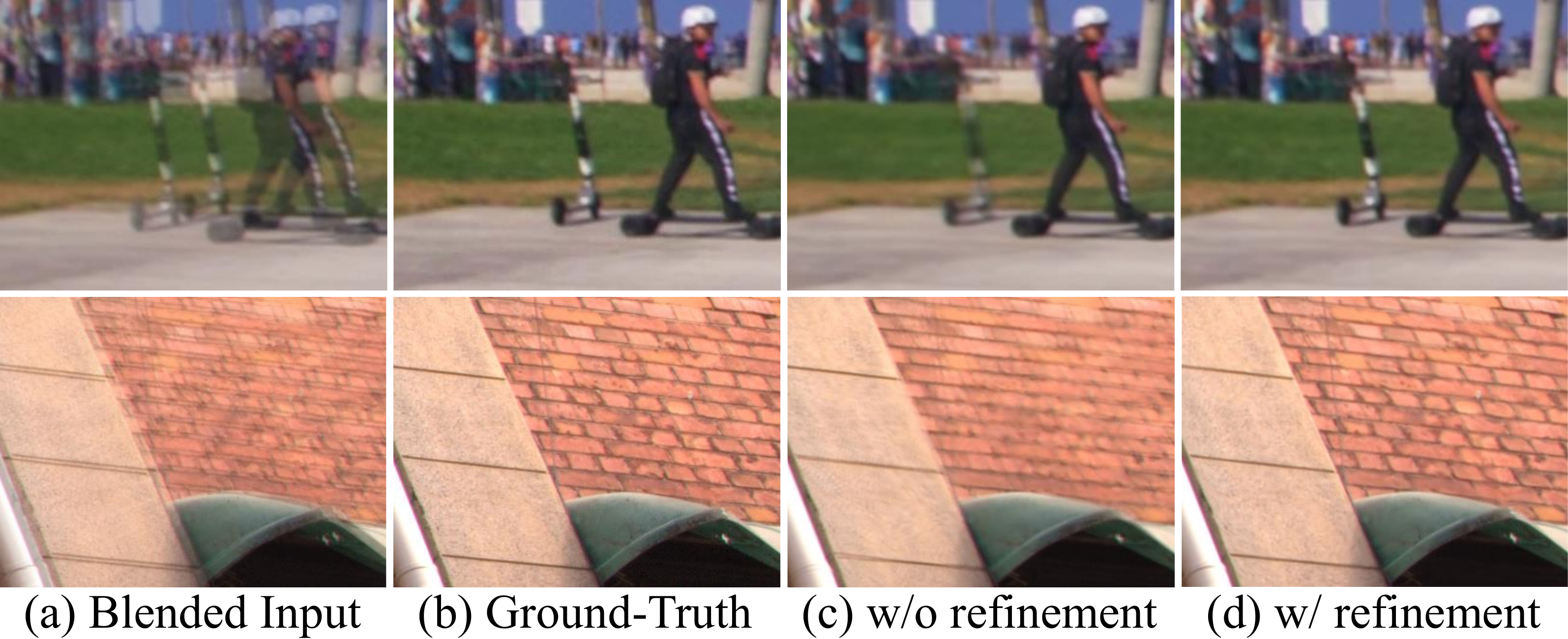}
    \vspace*{-0.5cm}
    \caption{
    Comparison of VFI results without and with the local motion refinement.}
    \label{fig:GMLM_ablation}
    \vspace*{-0.5cm}
\end{figure}

Figure~\ref{fig:motion_comparison} illustrates that refined motion fields are more accurate than global motion fields. Also, Figure~\ref{fig:GMLM_ablation} shows that  the local motion refinement improves VFI results by correcting errors in the global motion estimation. We see that motion errors on complicated texture and small objects in Figure~\ref{fig:GMLM_ablation}(c) are reduced by the local motion refinement in Figure~\ref{fig:GMLM_ablation}(d).

\begin{table}[t]
    \caption
    {
        Complexity analysis of the proposed algorithm.
    }
    \vspace*{-0.2cm}
    \footnotesize
    \centering
    \begin{tabular}{L{1.4cm}C{1.1cm}C{1.5cm}C{1.2cm}C{0.8cm}}
    \toprule
    & BiFormer & Upsampler & Synthesis & Total\\
    \midrule
    {\makecell{\footnotesize \#Parameters \\\footnotesize (millions)}}& 9.68& 0.78& 0.72 & 11.2\\
    \midrule
    {\makecell{\footnotesize 4K Runtime \\\footnotesize (seconds)}} & 0.02& 0.47& 1.74 & 2.23\\
    \bottomrule\\[-1em]
    \end{tabular}
    \label{table:complexity}
    \vspace*{-0.4cm}
\end{table}

\vspace*{0.2cm}
\noindent \textbf{Computational complexity}:
Table \ref{table:complexity} analyzes the complexity of the proposed algorithm, which consists of BiFormer, upsampler, and synthesis network. BiFormer uses 86\%  percent of parameters in the entire network. However, it operates at $1/8$ scale, so it takes only $0.02$ seconds to estimate global motion fields. On the other hand, the upsampler works at $1/4$ and $1/2$ scales and requires a longer processing time, although it is implemented efficiently with only 0.78M parameters. The frame synthesis network demands the longest processing time to synthesize a 4K frame. It is implemented with 0.72M parameters.

\section{Conclusions}

We proposed an effective 4K VFI algorithm based on BiFormer. First, we developed BiFormer to estimate global motion fields at a coarse scale. Second, we employed the upsampler to refine these global motion fields into final motion fields at a fine scale. Last, we used the synthesis network to warp the two input frames using the final motion fields, respectively, and synthesize an intermediate frame. It was shown that the proposed BiFormer algorithm provides excellent performance on 4K benchmark datasets.

\section*{Acknowledgments}
This work was conducted by CARAI grant funded by DAPA and ADD (UD190031RD), supported partly by Samsung Electronics Co., Ltd.~(No.~IO201214-08156-01), and supported partly by the NRF grants funded by the Korea government (MSIT) (No.~NRF-2021R1A4A1031864 and No.~NRF-2022R1A2B5B03002310).

\clearpage

{\small
\bibliographystyle{ieeetr}
\bibliography{egbib}
}

\end{document}